%% file: Main.tex
\documentclass[runningheads,a4paper]{llncs}

\usepackage{amssymb}
\usepackage{graphicx}
\usepackage[misc]{ifsym}
\usepackage{url}
\usepackage{amsmath,amssymb,amsfonts}
\usepackage{graphicx}
\usepackage{textcomp}
\usepackage{CJKutf8}
\usepackage{subfigure}
\usepackage{tabularx}
\usepackage{booktabs}
\usepackage{multirow}
\usepackage{xcolor}
\usepackage{wrapfig}
\usepackage{float}
\usepackage{threeparttable}      
\usepackage{nicefrac}      
\usepackage{microtype}      
\usepackage[utf8]{inputenc} 
\usepackage{hyperref}       
\usepackage{algpseudocode}
\usepackage{amsmath}

\usepackage{algorithm2e}
\usepackage{algpseudocode}

\urldef{\mailsa}\path|{zhuhongyu, jinxin4, liaohongcao}@ctbu.edu.cn|
\newcommand{\keywords}[1]{\par\addvspace\baselineskip
\noindent\keywordname\enspace\ignorespaces#1}

\begin{document}

\mainmatter  

\title{Relax DARTS: Relaxing the Constraints of Differentiable Architecture Search for Eye Movement Recognition}

\titlerunning{Relax DARTS}


\author{Hongyu Zhu$^1$$^\star$
\and Xin Jin$^1$\thanks{Equal contribution;  \Letter 
 Corresponding author}
\and Hongchao Liao$^1$  \and Yan Xiang$^3$  \and \\
Mounim~A.~El-Yacoubi$^2$
\and Huafeng Qin$^1$\Letter}

\authorrunning{Hongyu Zhu et al.}

%

\institute{$^1$Chongqing Technology and Business University,\\ $^2$Telecom SudParis, Institute Polytechnique de Paris,\\
$^3$Chongqing University of Arts and Sciences\\
\mailsa\\
}

%
%

\maketitle

\input{0_abstract}

\input{1_introduction}

\input{2_related_work}

\input{3_methodogy}

\input{4_Experiment}

\input{5_Conclusion}
\bibliographystyle{unsrt}
\bibliography{TEX}

\end{document}

%% file: 0_abstract.tex
\begin{abstract}
Eye movement biometrics is a secure and innovative identification method. Deep learning methods have shown good performance, but their network architecture relies on manual design and combined priori knowledge. To address these issues, we introduce automated network search (NAS) algorithms to the field of eye movement recognition and present Relax DARTS, which is an improvement of the Differentiable Architecture Search (DARTS) to realize more efficient network search and training. 
The key idea is to circumvent the issue of weight sharing by independently training the architecture parameters $\alpha$ to achieve a more precise target architecture. Moreover, the introduction of module input weights $\beta$ allows cells the flexibility to select inputs, to alleviate the overfitting phenomenon and improve the model performance. 
Results on four public databases demonstrate that the Relax DARTS achieves state-of-the-art recognition performance. Notably, Relax DARTS exhibits adaptability to other multi-feature temporal classification tasks.
\keywords{Eye movement biometrics, Differentiable Architecture Search.}

\end{abstract}

%% file: 1_introduction.tex
\section{Introduction}

Biometrics technology for authentication or identification and has been the subject of increasing attention in recent times. 
In contrast to traditional static physiological characteristics\cite{face2018,fingerprint2022,iris2004}, behavioral biometric modalities such as gait\cite{gait2023}, eye movements\cite{EmMixformer2024}, and handwriting\cite{handwriting2023}, show natural liveness identification, so as to achieve higher security. 

Among various behavioral biometric modalities, eye movement biometric systems show notable advantages to support liveness detection\cite{Makowski2020BiometricIA} and spoof-resistant continuous authentication\cite{eberz2015preventing}. Also, it can be easily and seamlessly integrated with iris\cite{iris2004}, pupil\cite{pupil2013}, and other ocular physiological attributes for multimodal recognition. Eye movement biometrics have received considerable attention in the past two decades\cite{katsini2020role}.
\begin{figure*}[t]
    \centerline{\includegraphics[scale=0.6]{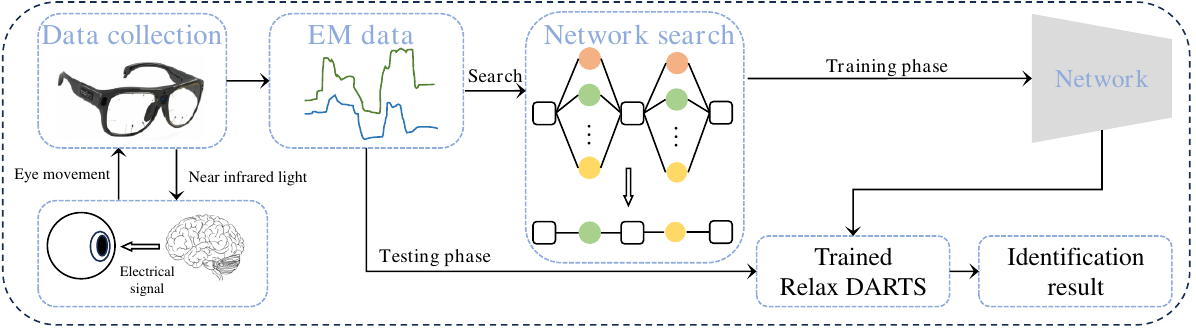}}
    \caption{The overall flow chart of eye movement (EM) recognition using the NAS algorithm. Compared to the existing DARTS-based algorithms\cite{DARTS2018,FairDarts2019,Chu2020DARTSR-}, we directly utilize the searched architectures for training and evaluation without stacking of units.}
    \label{fig1}
    \vspace{-10pt}
\end{figure*}

Eye movements are defined as the coordinated contraction and relaxation of the six ocular muscles, which are regulated by brainstem nerves during visual processing. These movements can be broadly classified into gaze and sweep based on the angular velocity of eye rotation. These movements permit the focusing of attention on objects of interest. Eye tracking devices employ near-infrared light reflections on the eye to detect the gaze point's position. This process captures the trajectory of gaze movements, thereby obtaining raw eye movement data. Furthermore, the resulting data undergoes feature coding and matching in order to achieve identification. As eye movements are under the control of the brain's neuroelectric signals, the data provides a wealth of information about the brain's cognitive processes occurring in real-time, and thus the eye movement is highly distinctive and stable over time\cite{Akkil2014TraQuMeAT}.

In this paper, to automatically search for models with excellent eye movement recognition performance, we propose Relax DARTS, a differentiable architecture search algorithm based on cell structure independent search with global search performance to relax the constraints of DARTS for eye movement recognition network search and training. Figure \ref{fig1} illustrates the general principle of Relax DARTS for achieving eye movement recognition. The key idea is to give independent search spaces to cells, increasing the degree of freedom during the network structure search by independently training each cell architecture parameter $\alpha$ in the network. Additionally, the input weight $\beta$ of the cell automatically selects cell input inputs.
The contributions of our work are summarized below:
\begin{itemize}
    \item We make the first attempt to introduce the NAS for eye movement recognition and propose a differentiable architecture search approach to automatically search powerful networks.
    \item The network has been relaxed to permit greater flexibility. The search strategy based on parameter sharing and cell stacking is discarded. In contrast, each cell is granted the autonomy to independently update the architectural parameters $\alpha$ and choose the operations.
    \item The network is granted global search capabilities. Each cell can choose the proportion of its inputs from direct and skip connections, or even None. Through the learning of the parameter $\beta$, enabling the network to global architectural adjustments while conducting local cell searches.
    \item We evaluated the Relax DARTS in terms of authentication performance in four public datasets.
    The experimental results demonstrated that our approach outperforms existing works in terms of reducing the verification error.

\end{itemize}

%% file: 2_related_work.tex
\section{Related Work}

\subsection{Traditional eye movement recognition method}
Traditional methods typically use manual techniques to filter useful features from eye movement data and then apply ML algorithms for feature extraction and recognition. Common features include eye movement velocity, gaze duration, and path length. 

In 2004, Kasprowski\cite{Kasprowski2004EyeMI} proved that eye movement data contains information that can be used for identity recognition firstly. In 2008, Komogortsev\cite{Komogortsev2008EyeMP} fully considered the bioanatomic properties of the eyeball and established a linear horizontal oculomotor plant mechanical model, \cite{Komogortsev2012BiometricAV} expanded on this foundation and proposed the concept of oculomotor plant characteristics (OPC). In 2017, Bayat\cite{Bayat2017BiometricIT} conducted recognition experiments on a self-constructed dataset, using eye movement data combined with pupil size. In 2018, Li\cite{Li2018BiometricRV} and colleagues extracted eye movement features using a multi-channel Gabor wavelet transform.

\subsection{Deep Learning based eye movement recognition method}
DL models are extensively utilized in computer vision (CV)\cite{Jin2024StarLKNetSM} and natural language processing (NLP)\cite{Subakan2020AttentionIA}, but also in the field of eye movement recognition, due to their capacity for end-to-end feature learning.

In 2019, Lena\cite{Jger2019DeepEB}developed a convolutional network (CNN)-based Siamese Network that utilizes eye movement data as input into two separate sub-networks for recognition.
These studies\cite{Makowski2021DeepEyedentificationLiveOB} fine-tuned the model and expanded it into the DeepEyedentificationLive (DEL) model. In 2022, Dillon et al.\cite{Lohr2022EyeKY} proposed exponentially-dilated CNNs for recognizing eye movement features.
In addition, in 2023, Taha\cite{Taha2023EyeDriveAD} et al. collected vehicle driver eye movement data via a remote low-frequency acquisition device, and extracted data features by combining Long Short Term Memory (LSTM) and dense networks to achieve end-to-end driver identification. In the same year, Qin et al.\cite{EmMixformer2024} extracted temporal and spatial features of eye-movement data for the combination using improved LSTM and Transformer algorithms to achieve state-of-the-art performance.

\subsection{Differentiable architecture search}
DARTS\cite{DARTS2018} uses $Softmax$ to treat the selection of candidate operations as an optimization problem for the architecture weight parameters in a continuous space so that the whole architecture search process can be differentiated to achieve gradient-based optimization, which is the most popular among the NAS algorithms at present.
DARTS+\cite{Liang2019DARTSID} proposes introducing the early stopping mechanism in the search stage to address the issue of skip-connection enrichment during the training process of DARTS. This issue leads to a significant performance loss of the final model. Fair DARTS \cite{FairDarts2019} also addresses the skip-connection enrichment phenomenon by transforming the candidate operations in the search phase from competition to cooperation. This is achieved by utilizing the Sigmoid function instead of Softmax to score the architectures. In addition, DARTS-\cite{Chu2020DARTSR-} proposed using auxiliary skip-connections to leverage the advantages of skip-connections in search operations compared to other operations.

%% file: 3_methodogy.tex
\begin{figure*}[t]
    \centerline{\includegraphics[scale=0.52]{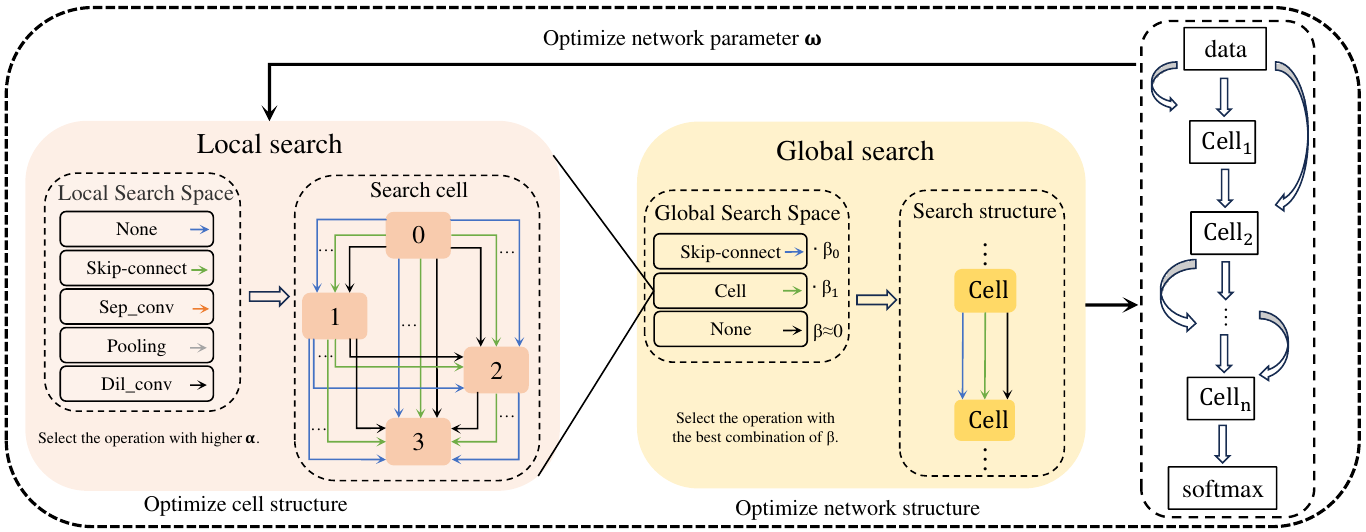}}
    \caption{The framework of the proposed Relax DARTS algorithm}
    \label{fig2}
\end{figure*}
\section{Approach}
This section presents the Relax DARTS algorithm for eye movement recognition. As shown in Fig.\ref{fig2},  our approach consists of two stages e.g. local cell search and global network search. As all cells do not share the weights, the representation capacity of the network is increased. Besides, our method reduces the gap between the proxy network and the target network by global search.

\subsection{Review Relax DARTS}
The objective of DARTS is to find a Normal Cell that maintains the output feature dimension and a Reduction Cell that reduces the output feature dimension by half. This is achieved by searching for the optimal combination of operations from the candidate operations in the search space.
In the search phase, Cell can be viewed as a directed acyclic graph. The path from node $i$ to $j$ represents a candidate operation $o^{(i,j)}\in \mathcal{O}$, and node $j$ represents a potential feature mapping $x^{(j)}$. Each Cell comprises two input nodes, four intermediate nodes $x^{(j)} = \sum_{i<j} o^{(i, j)}(x^{(i)})$, and one output node. 
DARTS assigns an architectural weight vector $\alpha^{(i,j)}$ to each $o^{(i,j)}$, and relaxes the categorical choice of a particular operation by the output probability distribution of $Softmax$:
\begin{align}
	Cell(x) = \bar{o}^{(i,j)}(x) = \sum_{o \in \mathcal{O}} \frac{\exp(\alpha_o^{(i,j)})}{\sum_{o' \in \mathcal{O}} \exp(\alpha_{o'}^{(i,j)})} o(x).
	\label{eq1}
\end{align}
The task of Cell architecture search then reduces to learning a matrix $\alpha$ consisting of a set of continuous variables $\{\alpha^{(i,j)}\}$.

In contrast to the $\alpha$ weight-sharing strategy in DARTS, each Cell in Relax DARTS performs select operations based on a distinct weight matrix $\alpha$. Furthermore, the ratio of the two input nodes in the $Cell_i$, namely the inputs from $Cell_{i-1}$ and $Cell_{i-2}$, is adjusted by setting the learnable weight parameter $\beta_i=\{{\beta_0}^i, {\beta_1}^i\}$. The inputs $s_0^i$ and $s_1^i$ of $Cell_i$ are represented as:
\begin{align}
	s_0^i = \frac{\exp({\beta_0}^i)}{\exp({\beta_0}^i)+\exp({\beta_1}^i)} Cell_{i-2}(x) \label{2}, \\
 s_1^i = \frac{\exp({\beta_1}^i)}{\exp({\beta_0}^i)+\exp({\beta_1}^i)} Cell_{i-1}(x).
	\label{3} 
\end{align}

The objective of the search phase is to optimize the weights $\alpha$, $\beta$, and $w$ in the supernet, which is formed by all the mixing operations, using gradient descent. The training and validation loss, denoted by $\mathcal{L}_{train}$ and $\mathcal{L}_{val}$, respectively, are determined by the weights $\alpha$, $\beta$ and $w$ together. Specifically, architecture search aims to find $\alpha^*$ and $\beta^*$ corresponding to minimizing $\mathcal{L}_{val}$ by Eq.(\ref{eq4} and \ref{eq5}) and the corresponding network weights $w^*$ by minimizing $\mathcal{L}_{train}$ by Eq.(\ref{eq6}). This task can be regarded as a triple optimization problem for joint optimization:
\begin{equation}
\alpha^* = \min\mathcal{L}_{val}(w^*(\alpha), \alpha) \label{eq4}, 
\end{equation}
\begin{equation}
\beta^* = \min \mathcal{L}_{val}(w^*(\alpha,\beta), \beta) \label{eq5}, 
\end{equation}
\begin{equation}
w^*(\alpha,\beta) = \mathrm{argmin}_w \enskip \mathcal{L}_{train}(w, \alpha^*, \beta^*). \label{eq6}
\end{equation}
After obtaining the optimal cell structure by searching, training on the target task is restarted.

\subsection{Local Search Strategy}
To minimize the performance loss caused by parameter $\alpha$ sharing and the stacking of identical cells in the network structure search, the local search strategy enables each cell to be endowed with an independent architectural parameter, thereby ensuring that each cell in the network is unique according to its current position. Specifically, we propose allowing each cell to update the structural parameter $\alpha$ independently. After initialization, $\alpha$ is used as the weight matrix for each cell's node operation, enabling independent architecture selection. 
To reduce the risk of overfitting, we simplify the network structure by alternately stacking the Normal Cell and Reduction Cell 3 times to become a 6-layer network. At the end of the search phase, we obtained six distinct cells instead of two.

\subsection{Global Search Strategy}
The performance gap that arises when using the searched network structure for training and evaluation has been a persistent issue due to differences in network architecture\cite{Cai2018ProxylessNASDN,Chen2019ProgressiveDA}. To address this, we introduce the learnable parameter $\beta$ as the input weight of each cell to fine-tune the network architecture during training and optimize the global network structure. The $cell_i$ has two input feature vectors, $s^i_0$ and $s^i_1$. The input $s^i_0$ is the same as the input $s^{i-1}_1$ of the previous $cell_{i-1}$, which is a skip-connect operation at the network structure level. The output of the previous cell is the other input $s^i_1$, which is a direct-connect operation.
The parameter $\beta$ is learnable and measures the proportion of cells that choose skip-connect and direct-connect operations. We use softmax for normalization after initializing $\beta$ in the same way as $\alpha$. The competition between the two inputs is preserved, and note that the proportions sum to 2 instead of 1, avoiding simultaneous scaling of the inputs. When the weights of the inputs are less than the threshold $c = 0.2$, the inputs are replaced with the all-zero tensor.

A network structure can be determined by a set of parameters $\alpha, \beta$, and $\omega$. Evaluating the performance of the structure after optimizing the parameters at each step is costly. Therefore, we use the same approximation strategy as DARTS to perform an alternating triple optimization of $\alpha, \beta$ and $\omega$ without training each network to convergence:

\begin{equation}
\nabla_\alpha \mathcal{L}_{val}(w, \alpha)
\approx \nabla_\alpha \mathcal{L}_{val}(w - \xi^\alpha \nabla_{w} \mathcal{L}_{train}(w, \alpha), \alpha),
\label{eq12}
\end{equation}
\begin{equation}
\nabla_\beta \mathcal{L}_{val}(w, \beta) 
\approx \nabla_\beta \mathcal{L}_{val}(w - \xi^\beta \nabla_{w} \mathcal{L}_{train}(w, \beta), \beta), \label{eq13}
\end{equation}
\begin{equation}
\nabla_w \mathcal{L}_{train}(w, (\alpha, \beta))\ \label{eq14},
\end{equation}
The parameter optimization learning rate is represented by $\xi$, and Algorithm\ref{alg1} illustrates the overall search algorithm.
\input{Algorithm1}

%% file: Algorithm1.tex
\RestyleAlgo{ruled}

\begin{algorithm*}
\caption{Relax DARTS}\label{alg1}
\KwIn{The $train$ and $val$ data; Architecture weights $\alpha$; Network weights $\omega$; Input weight $\beta$; Search epochs $E$;}
\KwOut{The best performing network structure}

Initialize $\alpha$ and $\beta$\;

Create a mixed operation parametrized by $\alpha$ for each edge and create a mixed input parametrized by $\beta$ for each cell\; 
    
Construct a supernet and initialize supernet weights $\omega$\;

\For{ each $\in$ [1, E] }
    {
         Sample $batch$ $\in$ $val$\;
         $Cell_i$ independently updates the $\alpha_i$ by $\nabla_\alpha\mathcal{L}_{val}$ and
         the $\beta_i$ by $\nabla_\beta\mathcal{L}_{val}$\;
         Sample $batch$ $\in$ $train$\;
         Update network weights $\omega$ by descending $\nabla_\omega\mathcal{L}_{train}$\;
         Derive the final architecture based on the learned $\alpha$ and $\beta$\;
    }
\end{algorithm*}

%% file: 4_Experiment.tex
\section{Experiments}
To evaluate our approach, we performed extensive experiments on four public datasets:  JuDo1000\cite{Makowski2020BiometricIA} and three sub-datasets (RAN, HSS, and TEX) in GazeBase \cite{Griffith2021GazeBaseAL}. We compare Relax DARTS with state-of-the-art work, including not only eye movement recognition algorithms: the DEL \cite{Jger2019DeepEB}, Expansion CNN\cite{Lohr2021EyeKY}, Dense LSTM\cite{Taha2023EyeDriveAD}, EKTY\cite{Lohr2022EyeKY} and EmMixformer\cite{EmMixformer2024}, but also the classical DARTS\cite{DARTS2018} algorithm as well as its improvements Fair-DARTS\cite{FairDarts2019} and DARTS-\cite{Chu2020DARTSR-}. 

The network is partitioned and trained directly on the dataset to be evaluated, without a proxy dataset, because the computational overhead is affordable and reduces the performance gap between search and evaluation.

All DARTS-based methods perform 50 search epochs and 300 training epochs, following the same network parameter settings as in \cite{FairDarts2019} and the same eye movement data partitioning as in \cite{EmMixformer2024} for fair baseline testing and comparison. We set the batch size to 32 for training and 256 for testing, the learning rate decays from 0.025 to 0 with cosine annealing, SGD as the optimizer with a momentum of 0.9 and a weight decay of $5 \times
10^-4$, and the drop-path rate is set to 0.3. All experiments were done in PyTorch and implemented on a high-performance computer with an NVIDIA A100 Tensor Core GPU.
\begin{figure*}[b]
    \begin{minipage}{0.5\linewidth}
        \input{table_RAN}
    \end{minipage}
    \begin{minipage}{0.5\linewidth}
        \input{table_HSS}
    \end{minipage}
\end{figure*}

\subsection{Identification Performance of Relax DARTS}
In detail, in the traditional deep learning approach, we select the first round of data for training and the second round of data collected at different times for testing. In the DARTS-based approach, the first round of data is first divided into a training set and a validation set for the automatic search of the network, and the resulting target network architecture is then utilized for training with the first round of data and testing with the second round of data. 

Tables \ref{table_ran}-\ref{table_tex} list the EER and FRR@FAR of each approach. It is evident that our Relax DARTS outperforms existing approaches, achieving the lowest verification error, i.e. 0.0657, 0.0610, and 0.0529 on the three subsets of the GazeBase dataset, and 0.0437 on the JuDo1000 dataset. Also, it is clear that our proposed approach achieves higher recognition accuracy compared to existing approaches at different FARs.
\begin{figure*}[t]
    \vspace{-20pt}
    \begin{minipage}{0.5\linewidth}
        \input{table_TEX}
    \end{minipage}
    \begin{minipage}{0.5\linewidth}
        \input{table_JUDO}
    \end{minipage}
    \vspace{-20pt}
\end{figure*}

\subsection{Ablation Experiment}
To investigate the effect of each step on the model's recognition accuracy improvement, we performed ablation experiments on the RAN sub-dataset in GazeBase. Specifically, we used DARTS as a baseline and added $\alpha$-independent optimization strategy, resulting in the model denoted as ‘+$\alpha$’. Finally, we introduced cell input weights $\beta$ for global network structure fine-tuning, resulting in the model expressed as ‘+$\beta$(Relax DARTS)’.

The verification error rates resulting from the ablation schemes are presented in Table\ref{table_ablation}. The experimental results suggest that $\alpha$-independent optimization can significantly enhance the baseline model's recognition performance. This is because the open $\alpha$-optimization strategy increases the degree of freedom of network search, resulting in unique cells that are adapted to the current network location. In addition, including input weights $\beta$ achieves the lowest EER and yields the best performance. This enables fine-tuning of the network architecture while the network performs cell searching, allowing for global search.
\input{ablation}

%% file: table_RAN.tex
\begin{table}[H]
\vspace{-35pt}
\caption{Results of comparative experiments on the RAN database}
\centering
\resizebox{1.0\linewidth}{!}{
    \begin{tabular}{ c c c c c }
    \toprule
        \multirow{2}*{RAN} &\multirow{2}*{EER} & \multicolumn{3}{ c }{FRR@FAR}\\
        \cline{3-5}
        &  & $10^{-1}$ & $10^{-2}$  & $10^{-3}$\\
        \hline
        \textbf{Relax DARTS}  &\textbf{0.0657} & \textbf{0.0452} & \textbf{0.2648} & \textbf{0.6681} \\
        
        DARTS \cite{DARTS2018} & 0.0956 & 0.0910 & 0.5412 & 0.8987 \\
        
        Fair DARTS\cite{FairDarts2019} & 0.0720  & 0.0532  & 0.3048 & 0.7038 \\
        
        DARTS-\cite{Chu2020DARTSR-} & 0.0788 & 0.0617 & 0.3907 & 0.7915  \\
        
        EmMixFormer\cite{EmMixformer2024} & 0.0801 & 0.0680 & 0.2818 & 0.2818\\
        
        DEL\cite{Makowski2021DeepEyedentificationLiveOB} & 0.1436 & 0.2066 & 0.7383 & 0.9645\\
        
        Expansion CNN\cite{Lohr2021EyeKY} & 0.1500 & 0.2340 & 0.7277 & 1.0000\\
        
        Dense LSTM\cite{Taha2023EyeDriveAD} & 0.1161 & 0.1329 & 0.5529 & 0.8846\\
        
        EKYT\cite{Lohr2022EyeKY}& 0.0885 & 0.0807 & 0.3513 & 0.7045\\
    \bottomrule
    \end{tabular}
}
\label{table_ran}
\end{table}

%% file: table_HSS.tex
\begin{table}[H]
\vspace{-35pt}
\caption{Results of comparative experiments on the HSS database}
\centering
\resizebox{1.0\linewidth}{!}{
    \begin{tabular}{ c c c c c }
    \toprule
        \multirow{2}*{HSS} &\multirow{2}*{EER} & \multicolumn{3}{c}{FRR@FAR}\\
        \cline{3-5}
        &  & $10^{-1}$ & $10^{-2}$  & $10^{-3}$\\
        \hline
        \textbf{Relax DARTS}  & \textbf{0.0610} & \textbf{0.0398} & \textbf{0.1504} & \textbf{0.3718}\\
        
        DARTS \cite{DARTS2018} & 0.0642 & 0.0423 & 0.2307 & 0.5448\\
        
        Fair DARTS\cite{FairDarts2019} & 0.0736 & 0.0544 & 0.2939 & 0.6424\\
        
        DARTS-\cite{Chu2020DARTSR-} & 0.0686 & 0.0482 & 0.2634 & 0.6119\\
        
        EmMixFormer\cite{EmMixformer2024} & 0.0673 &0.0502 & 0.2032 & 0.4659\\
        
        DEL\cite{Makowski2021DeepEyedentificationLiveOB} &  0.1309 & 0.1680 & 0.6217 & 0.9087\\
        
        Expansion CNN\cite{Lohr2021EyeKY} & 0.1437 & 0.1894 & 0.6267 & 1.0000\\
        
        Dense LSTM\cite{Taha2023EyeDriveAD} & 0.1191 &0.1365 & 0.5082 & 0.8382\\
        
        EKYT\cite{Lohr2022EyeKY}& 0.0839 & 0.0739 & 0.2691 & 0.5575\\
    \bottomrule
    \end{tabular}
}
\label{table_hss}
\end{table}

%% file: table_TEX.tex
\begin{table}[H]
\caption{Results of comparative experiments on the TEX database}
\centering
\resizebox{1.0\linewidth}{!}{
\begin{tabular}{ c c c c c }
\toprule
\multirow{2}*{TEX} &\multirow{2}*{EER} & \multicolumn{3}{ c }{FRR@FAR}\\
\cline{3-5}
&  & $10^{-1}$ & $10^{-2}$  & $10^{-3}$\\
\hline
\textbf{Relax DARTS}  & \textbf{0.0529} & \textbf{0.0324} & \textbf{0.1362} & \textbf{0.4596}\\

DARTS \cite{DARTS2018} & 0.0580 & 0.0338 & 0.2160 & 0.5606\\

Fair DARTS\cite{FairDarts2019} & 0.0583 & 0.0347 & 0.2183 & 0.5445 \\

DARTS-\cite{Chu2020DARTSR-} & 0.0596 & 0.0343 & 0.2304 & 0.5717 \\

EmMixFormer\cite{EmMixformer2024} & 0.0635 & 0.0407 & 0.2603 & 0.6193\\

DEL\cite{Makowski2021DeepEyedentificationLiveOB} & 0.1060 & 0.1128 & 0.5750 & 0.9141\\

Expansion CNN\cite{Lohr2021EyeKY}& 0.1362 & 0.1950 & 0.6977 & 1.0000 \\

Dense LSTM\cite{Taha2023EyeDriveAD}& 0.0971 & 0.0945 & 0.4824 & 0.8456 \\

EKYT\cite{Lohr2022EyeKY}& 0.0736 & 0.0551 & 0.3293  & 0.7175 \\
\bottomrule
\end{tabular}
}
\label{table_tex}
\end{table}

%% file: table_JUDO.tex
\begin{table}[H]
\caption{Results of comparative experiments on the JuDo1000 database}
\centering
\resizebox{1.0\linewidth}{!}{
    \begin{tabular}{ c c c c c }
    \toprule
    \multirow{2}*{JuDo1000} &\multirow{2}*{EER} & \multicolumn{3}{ c }{FRR@FAR}\\
    \cline{3-5}
    &  & $10^{-1}$ & $10^{-2}$  & $10^{-3}$\\
    \hline
    \textbf{Relax DARTS}  & \textbf{0.0437} & \textbf{0.0204} & \textbf{0.0953} & \textbf{0.3296} \\
    
    DARTS \cite{Liu2018DARTSDA} & 0.0583 & 0.0339 & 0.2035 & 0.4918\\
    
    Fair DARTS\cite{Chu2019FairDE} & 0.0588 & 0.0365 &0.1953  & 0.4848 \\
    
    DARTS-\cite{Chu2020DARTSRS} & 0.0577 & 0.0357 & 0.1911 & 0.4772 \\
    
    EmMixFormer\cite{Qin2024EmMixformerMT} &0.0543 & 0.0059 & 0.1284 & 0.3359 \\
    
    DEL\cite{Makowski2021DeepEyedentificationLiveOB} & 0.1238 & 0.0781  &  0.5508 & 0.8945 \\
    
    Expansion CNN\cite{Lohr2021EyeKY}  & 0.0989 & 0.0586 & 0.3203 & 0.7594 \\
    
    Dense LSTM\cite{Taha2023EyeDriveAD} & 0.0669 & 0.0195 & 0.2305 & 0.6016\\
    
    EKYT\cite{Lohr2022EyeKY}& 0.0773 & 0.0125 & 0.1953  & 0.4922 \\
    \bottomrule
    \end{tabular}
}
\label{table_JUDO}
\end{table}

%% file: ablation.tex
\begin{table}[!htbp]
\caption{Results of the ablation experiments on the RAN database}
\centering
\begin{tabular}{ c c c c c }
\toprule
\multirow{2}*{RAN} &\multirow{2}*{EER} & \multicolumn{3}{ c }{FRR@FAR}\\
\cline{3-5}
&  & $10^{-1}$ & $10^{-2}$  & $10^{-3}$\\
\hline
DARTS \cite{DARTS2018} & 0.0956 & 0.0910 & 0.5412 & 0.8987 \\
$+\alpha$  & 0.0777 & 0.0599 & 0.3648 & 0.7654\\
\textbf{$+\beta$ (Relax DARTS)}  & \textbf{0.0657} & \textbf{0.0452} & \textbf{0.2648} & \textbf{0.6681} \\

\bottomrule
\end{tabular}
\label{table_ablation}
\end{table}

%% file: 5_Conclusion.tex
\section{Conclusion}
In this paper, we relax the search strategy of DARTS and propose Relax DARTS, which implements alternating local and global architectures for end-to-end model search and training for eye movement biometric authentication. 
Our experimental results demonstrate that our approach outperforms existing methods and achieves a new state-of-the-art verification accuracy.

%% file: Main.bbl
\begin{thebibliography}{10}

\bibitem{face2018}
Yu~Liu, Fangyin Wei, Jing Shao, Lu~Sheng, Junjie Yan, and Xiaogang Wang.
\newblock Exploring disentangled feature representation beyond face identification.
\newblock {\em 2018 IEEE/CVF Conference on Computer Vision and Pattern Recognition}, pages 2080--2089, 2018.

\bibitem{fingerprint2022}
Yang Peng, Peng Liu, Yu~Wang, Guan Gui, Bamidele Adebcisi, and Haris Gacanin.
\newblock Radio frequency fingerprint identification based on slice integration cooperation and heat constellation trace figure.
\newblock {\em IEEE Wireless Communications Letters}, 11:543--547, 2022.

\bibitem{iris2004}
J.~Daugman.
\newblock How iris recognition works.
\newblock {\em IEEE Transactions on Circuits and Systems for Video Technology}, 14(1):21--30, 2004.

\bibitem{gait2023}
Chi Xu, Yasushi Makihara, Xiang Li, and Yasushi Yagi.
\newblock Occlusion-aware human mesh model-based gait recognition.
\newblock {\em IEEE Transactions on Information Forensics and Security}, 18:1309--1321, 2023.

\bibitem{EmMixformer2024}
Huafeng Qin, Hongyu Zhu, Xin Jin, Qun Song, Moun{\^i}m~A. El-Yacoubi, and Xinbo Gao.
\newblock Emmixformer: Mix transformer for eye movement recognition.
\newblock {\em ArXiv}, abs/2401.04956, 2024.

\bibitem{handwriting2023}
Chun-Xia Yang, Dongzhi Zhang, Dongyue Wang, Huixin Luan, Xiaoya Chen, and Weiyu Yan.
\newblock In situ polymerized mxene/polypyrrole/hydroxyethyl cellulose-based flexible strain sensor enabled by machine learning for handwriting recognition.
\newblock {\em ACS applied materials \& interfaces}, 2023.

\bibitem{Makowski2020BiometricIA}
Silvia Makowski, Lena~A. J{\"a}ger, Paul Prasse, and Tobias Scheffer.
\newblock Biometric identification and presentation-attack detection using micro- and macro-movements of the eyes.
\newblock {\em 2020 IEEE International Joint Conference on Biometrics (IJCB)}, pages 1--10, 2020.

\bibitem{eberz2015preventing}
Simon Eberz, Kasper Rasmussen, Vincent Lenders, and Ivan Martinovic.
\newblock Preventing lunchtime attacks: Fighting insider threats with eye movement biometrics.
\newblock In {\em Network and Distributed System Security (NDSS) Symposium}. Internet Society, 2015.

\bibitem{pupil2013}
Pieter~J. Blignaut.
\newblock Mapping the pupil-glint vector to gaze coordinates in a simple video-based eye tracker.
\newblock {\em Journal of Eye Movement Research}, 7, 2013.

\bibitem{katsini2020role}
Christina Katsini, Yasmeen Abdrabou, George~E Raptis, Mohamed Khamis, and Florian Alt.
\newblock The role of eye gaze in security and privacy applications: Survey and future hci research directions.
\newblock In {\em Proceedings of the 2020 CHI Conference on Human Factors in Computing Systems}, pages 1--21, 2020.

\bibitem{DARTS2018}
Hanxiao Liu, Karen Simonyan, and Yiming Yang.
\newblock Darts: Differentiable architecture search.
\newblock {\em International Conference on Learning Representations}, abs/1806.09055, 2019.

\bibitem{FairDarts2019}
Xiangxiang Chu, Tianbao Zhou, Bo~Zhang, and Jixiang Li.
\newblock Fair darts: Eliminating unfair advantages in differentiable architecture search.
\newblock In {\em European conference on computer vision}, pages 465--480. Springer, 2020.

\bibitem{Chu2020DARTSR-}
Xiangxiang Chu, Xiaoxing Wang, Bo~Zhang, Shun Lu, Xiaolin Wei, and Junchi Yan.
\newblock Darts-: Robustly stepping out of performance collapse without indicators.
\newblock {\em International Conference on Learning Representations}, abs/2009.01027, 2021.

\bibitem{Akkil2014TraQuMeAT}
Deepak Akkil, Poika Isokoski, Jari Kangas, Jussi Rantala, and R.~Raisamo.
\newblock Traqume: a tool for measuring the gaze tracking quality.
\newblock {\em Proceedings of the Symposium on Eye Tracking Research and Applications}, 2014.

\bibitem{Kasprowski2004EyeMI}
Paweł Kasprowski and J{\'o}zef Ober.
\newblock Eye movements in biometrics.
\newblock In {\em ECCV Workshop BioAW}, 2004.

\bibitem{Komogortsev2008EyeMP}
Oleg~V. Komogortsev and Javed~I. Khan.
\newblock Eye movement prediction by kalman filter with integrated linear horizontal oculomotor plant mechanical model.
\newblock {\em Proceedings of the 2008 symposium on Eye tracking research \& applications}, 2008.

\bibitem{Komogortsev2012BiometricAV}
Oleg~V. Komogortsev, Alexey Karpov, Larry~R. Price, and Cecilia~R. Aragon.
\newblock Biometric authentication via oculomotor plant characteristics.
\newblock {\em 2012 5th IAPR International Conference on Biometrics (ICB)}, pages 413--420, 2012.

\bibitem{Bayat2017BiometricIT}
Akram Bayat and Marc Pomplun.
\newblock Biometric identification through eye-movement patterns.
\newblock In {\em International Conference on Applied Human Factors and Ergonomics}, 2017.

\bibitem{Li2018BiometricRV}
Chunyong Li, Jiguo Xue, Cheng Quan, Jingwei Yue, and Chenggang Zhang.
\newblock Biometric recognition via texture features of eye movement trajectories in a visual searching task.
\newblock {\em PLoS ONE}, 13, 2018.

\bibitem{Jin2024StarLKNetSM}
Xin Jin, Hongyu Zhu, Moun{\^i}m~A. El-Yacoubi, Hongchao Liao, Huafeng Qin, and Yun Jiang.
\newblock Starlknet: Star mixup with large kernel networks for palm vein identification.
\newblock {\em ArXiv}, abs/2405.12721, 2024.

\bibitem{Subakan2020AttentionIA}
Cem Subakan, Mirco Ravanelli, Samuele Cornell, Mirko Bronzi, and Jianyuan Zhong.
\newblock Attention is all you need in speech separation.
\newblock {\em ICASSP 2021 - 2021 IEEE International Conference on Acoustics, Speech and Signal Processing (ICASSP)}, pages 21--25, 2020.

\bibitem{Jger2019DeepEB}
Lena~A. J{\"a}ger, Silvia Makowski, Paul Prasse, Sascha Liehr, Maximilian Seidler, and Tobias Scheffer.
\newblock Deep eyedentification: Biometric identification using micro-movements of the eye.
\newblock {\em ArXiv}, abs/1906.11889, 2019.

\bibitem{Makowski2021DeepEyedentificationLiveOB}
Silvia Makowski, Paul Prasse, David~Robert Reich, Daniel~G. Krakowczyk, Lena~A. J{\"a}ger, and Tobias Scheffer.
\newblock Deepeyedentificationlive: Oculomotoric biometric identification and presentation-attack detection using deep neural networks.
\newblock {\em IEEE Transactions on Biometrics, Behavior, and Identity Science}, 3:506--518, 2021.

\bibitem{Lohr2022EyeKY}
Dillon~James Lohr and Oleg~V. Komogortsev.
\newblock Eye know you too: A densenet architecture for end-to-end biometric authentication via eye movements.
\newblock {\em ArXiv}, abs/2201.02110, 2022.

\bibitem{Taha2023EyeDriveAD}
Bilal Taha, Sherif Nagib~Abbas Seha, Dae~Yon Hwang, and Dimitrios Hatzinakos.
\newblock Eyedrive: A deep learning model for continuous driver authentication.
\newblock {\em IEEE Journal of Selected Topics in Signal Processing}, 17:637--647, 2023.

\bibitem{Liang2019DARTSID}
Hanwen Liang, Shifeng Zhang, Jiacheng Sun, Xingqiu He, Weiran Huang, Kechen Zhuang, and Zhenguo Li.
\newblock Darts+: Improved differentiable architecture search with early stopping.
\newblock {\em ArXiv}, abs/1909.06035, 2019.

\bibitem{Cai2018ProxylessNASDN}
Han Cai, Ligeng Zhu, and Song Han.
\newblock Proxylessnas: Direct neural architecture search on target task and hardware.
\newblock {\em ArXiv}, abs/1812.00332, 2018.

\bibitem{Chen2019ProgressiveDA}
Xin Chen, Lingxi Xie, Jun Wu, and Qi~Tian.
\newblock Progressive differentiable architecture search: Bridging the depth gap between search and evaluation.
\newblock {\em 2019 IEEE/CVF International Conference on Computer Vision (ICCV)}, pages 1294--1303, 2019.

\bibitem{Griffith2021GazeBaseAL}
Henry~K. Griffith, Dillon~James Lohr, Evgeny Abdulin, and Oleg~V. Komogortsev.
\newblock Gazebase, a large-scale, multi-stimulus, longitudinal eye movement dataset.
\newblock {\em Scientific Data}, 8, 2021.

\bibitem{Lohr2021EyeKY}
Dillon~James Lohr, Henry~K. Griffith, and Oleg~V. Komogortsev.
\newblock Eye know you: Metric learning for end-to-end biometric authentication using eye movements from a longitudinal dataset.
\newblock {\em IEEE Transactions on Biometrics, Behavior, and Identity Science}, 4:276--288, 2021.

\bibitem{Chu2019FairDE}
Xiangxiang Chu, Tianbao Zhou, Bo~Zhang, and Jixiang Li.
\newblock Fair darts: Eliminating unfair advantages in differentiable architecture search.
\newblock In {\em European conference on computer vision}, pages 465--480. Springer, 2020.

\end{thebibliography}
